\documentclass[journal]{IEEEtran}
\usepackage{amsmath,amsfonts}
\usepackage{algorithmic}
\usepackage{algorithm}
\usepackage{array}
\usepackage[caption=false,font=normalsize,labelfont=sf,textfont=sf]{subfig}
\usepackage{textcomp}
\usepackage{stfloats}
\usepackage{url}
\usepackage{verbatim}
\usepackage{graphicx}
\usepackage{cite}
\usepackage{amssymb}
\usepackage{color}
\usepackage{bm}
\usepackage{hyperref}
\usepackage{makecell}
\usepackage{multirow}
\usepackage{fancyhdr}
\newtheorem{theorem}{Theorem}

\newtheorem{assumption}{Assumption}
\newtheorem{remark}{Remark}
\newtheorem{definition}{Definition}
\fancypagestyle{firstpage}{
    \fancyhf{} 
    \fancyhead[C]{\textcolor{red}{This work has been submitted to the IEEE for possible publication. Copyright may be transferred without notice, after which this version may no longer be accessible.}} 
}

\begin{document}

\title{Equilibrium Adaptation-Based Control for Track Stand of Single-Track Two-Wheeled Robots}

\author{Boyi~Wang,
        Yang~Deng,
        Feilong~Jing,
        Yiyong~Sun,~\emph{Member,~IEEE},\\
        Zhang~Chen$^{*}$,~\emph{Member,~IEEE},
        and~Bin~Liang,~\emph{Senior Member,~IEEE}
\thanks{This work was supported in part by National Natural Science Foundation of China under Grants 62073183 and 62203252. (\textit{Corresponding author: Zhang Chen.})}
\thanks{Boyi Wang, Yang Deng, Feilong Jing, Zhang Chen, and Bin Liang are with the Department of Automation, Tsinghua University, Beijing 100084, China (email: by-wang19@mails.tsinghua.edu.cn; dengyang@mail.tsinghua.edu.cn; jfl21@mails.tsinghua.edu.cn; cz\_da@tsinghua.edu.cn; bliang@tsinghua.edu.cn.)}
\thanks{Yiyong Sun is with the School of Aerospace Engineering, Beijing Institute of Technology, Beijing 100081, China. (email: sunyy@bit.edu.cn)}
}

\maketitle

\thispagestyle{firstpage}

\begin{abstract}
Stationary balance control is challenging for single-track two-wheeled (STTW) robots due to the lack of elegant balancing mechanisms and the conflict between the limited attraction domain and external disturbances. To address the absence of balancing mechanisms, we draw inspiration from cyclists and leverage the track stand maneuver, which relies solely on steering and rear-wheel actuation. To achieve accurate tracking in the presence of matched and mismatched disturbances, we propose an equilibrium adaptation-based control (EABC) scheme that can be seamlessly integrated with standard disturbance observers and controllers. This scheme enables adaptation to slow-varying disturbances by utilizing a disturbed equilibrium estimator, effectively handling both matched and mismatched disturbances in a unified manner while ensuring accurate tracking with zero steady-state error. We integrate the EABC scheme with nonlinear model predictive control (MPC) for the track stand of STTW robots and validate its effectiveness through two experimental scenarios. Our method demonstrates significant improvements in tracking accuracy, reducing errors by several orders of magnitude.
\end{abstract}

\begin{IEEEkeywords}
Disturbance rejection, STTW robot, track stand, mismatched disturbance.
\end{IEEEkeywords}

\section{Introduction}\label{sec:introduction}

\IEEEPARstart{S}{ingle}-track two-wheeled (STTW) robots, represented by unmanned motorcycles and bicycles, have gained increasing attention from researchers due to their advantages of high speed, narrow body and exceptional maneuverability compared to multi-track robots, multi-wheel robots, and legged robots\cite{chen2022gaussian,yi2006trajectory,tian2022steady,sun2020polynomial,getz1994control,getz1995control}.
However, the stationary instability of STTW robots hinders them from performing complex tasks, such as those requiring intermittent stops, thus impeding their practical application~\cite{zhang2011balance,yu2018steering}. 

The challenges faced by stationary balance control can be categorized into two dimensions. Firstly, the balancing mechanism at high speeds becomes ineffective at low speeds, resulting in the requirement of additional actuators (\textit{e.g.} pendulum, reaction wheel, control moment gyro)~\cite{he2022learning,cui2020nonlinear,wang2017trajectory,wang2023observer}. Secondly, STTW robots are affected by both matched and mismatched disturbances~\cite{wang2023tieequilibrium, wang2023iciraequilibrium}, and existing disturbance rejection methods cannot effectively address these disturbances in nonlinear, non-minimum phase systems. 
Therefore, it is necessary to address two key problems: (i) Is there any elegant physical mechanism for stationary balance? (ii) How can the robot handle matched and mismatched disturbances while ensuring high tracking accuracy? 

\subsection{Mechanisms for Stationary Balance}

Balance control of STTW robots is mainly based on steering motion, where the fundamental principle involves controlling the centrifugal force generated by steering to counteract the destabilizing effect of gravity \cite{meijaard2007linearized, yu2018steering}. However, a low speed cannot generate sufficient centrifugal force, making balance control unattainable. To address this issue, previous studies have examined a variety of strategies, including mainly four techniques: pendulum \cite{he2022learning}, reaction wheel \cite{cui2020nonlinear}, control moment gyroscope (CMG) \cite{wang2017trajectory,wang2023observer}, and steering control with negative trail \cite{zhang2011balance,yu2018steering}. However, the CMG and pendulum schemes introduce additional mass and actuators, while the reaction wheel scheme is not suitable for larger vehicles, and the negative trail scheme has a significantly restricted attraction domain \cite{zhang2011balance}. As a result, none of these techniques achieves a graceful stationary balance.

\begin{figure*}[t]
\centering
    \includegraphics[width=\textwidth]{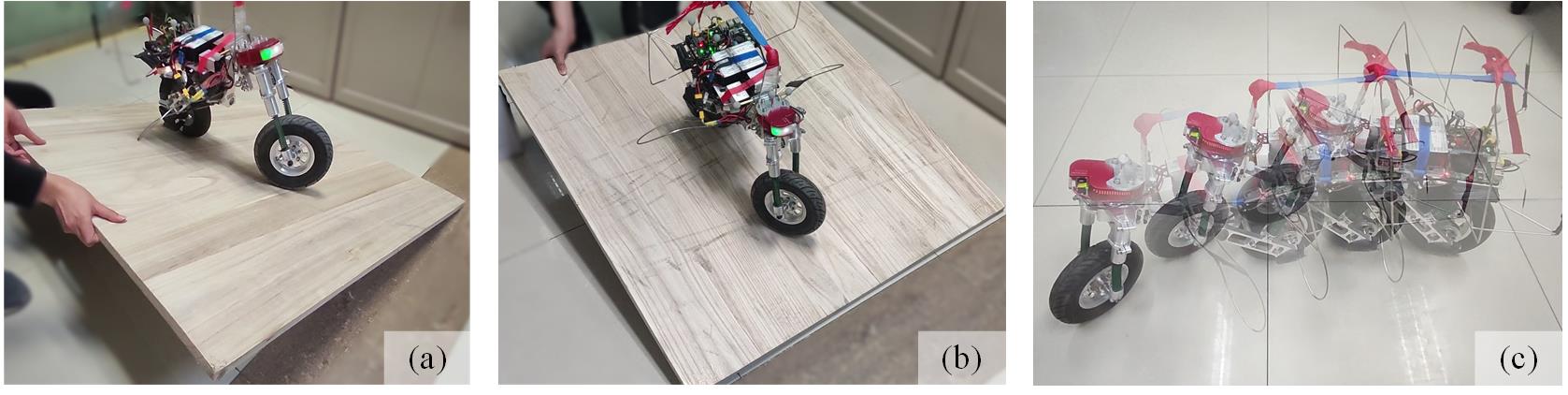}
    \caption{(a)/(b) The STTW robot is performing track stand on a laterally/longitudinally inclined plane with varying angles. (c) The STTW robot is tracking a rear position reference.}
    \label{fig:intro-all-exp-photos}
\end{figure*}

We have observed a unique stationary balance maneuver performed by human cyclists, known as the \textit{track stand}~\cite{wikipediaTrackStand,youtubeTrackStand}. This maneuver involves maintaining the handlebar at a specific angle and achieving balance through slight forward and backward movements of the rear wheel, as shown in Fig.~\ref{fig:intro-all-exp-photos}. Existing research has not well studied the track stand maneuver except for a series of studies by Huang et al.~\cite{huang2013BalancedMotions,huang2015simple}, which, however, only investigated the track stand maneuver of an uncommon structure with a motor-driven front wheel. The long-term disregard of this intriguing maneuver is partially due to the difficulty of modeling STTW robots. First, previous studies typically employed the Lagrangian method \cite{getz1994control,yi2006trajectory,tian2022steady,sun2020polynomial,getz1995control,yu2018steering,zhang2011balance}, resulting in a dynamic model lacking intuitive interpretation. Moreover, many studies directly controlled the velocity of the rear wheel \cite{he2015ConstantvelocitySteering,yu2018steering, wang2023iciraequilibrium}, rather than the torque, thus overlooking the effect of the rear wheel on balance control. 
To gain a more intuitive understanding of the principles behind track stand maneuver, we adopt the Newton-Euler method to establish the dynamics of a typical STTW robot equipped with a steering motor and a rear wheel motor. 
Through the equation of motion, we provide an interpretation for the mechanism of the track stand maneuver.

\subsection{Methods for Matched/Mismatched Disturbance Rejection}
For the stationary balance task, STTW robots face the challenge of matched and mismatched disturbances~\cite{chen2022gaussian,wang2023tieequilibrium}. Our goal is to design a disturbance rejection scheme to achieve accurate tracking and disturbance rejection simultaneously.

Compared to robust control methods such as $H_2/H_\infty$ control and sliding mode control (SMC), disturbance observer-based control (DOBC) has the advantages of flexibility and low conservatism~\cite{chen2011NonlinearDisturbance}, which has gained widespread applications, such as robotics~\cite{chen2000nonlinear, chen2004disturbance}, motor drives~\cite{yang2012NonlinearDisturbance,yang2013continuous,yang2017DisturbanceUncertainty}, hypersonic vehicles~\cite{sun2014NonlinearDisturbance}, mechatronic systems~\cite{li2012GeneralizedExtended,yang2013SlidingmodeControl}, and power systems~\cite{sun2015GlobalOutput}. The core idea of DOBC is using a disturbance observer (DO) to estimate disturbances and then introduce an equal and opposite quantity in the control input to counteract them, that is, ${u} = {\alpha(\bm{x})} - \hat{{d}}$, where, ${\alpha}(\bm{x})$ denotes a nominal feedback control law and $\hat{{d}}$ represents an estimated disturbance. However, this approach is only effective for matched disturbances.

For mismatched disturbances, Chen et al. \cite{chen2011NonlinearDisturbance} proposed nonlinear disturbance observer-based robust control (NDOBRC) for nonlinear single-input single-output (SISO) systems with mismatched disturbances, which introduced a disturbance compensation gain vector into the control law, in the form ${u} = {\alpha}(\bm{x}) + {\beta}(\bm{x})\hat{{d}}$, to eliminate mismatched disturbances from the output channel. However, the calculation of the gain ${\beta}(\bm{x})$ is complex and is coupled with the nominal control law ${\alpha}(\bm{x})$.
Yang et al. \cite{yang2012NonlinearDisturbance} developed a generalized DOBC method for multiple-input multiple-output (MIMO) nonlinear systems, where the systems are assumed to be capable of state feedback linearization.
Meanwhile, in various control fields, researchers have developed method-specific techniques for mismatched disturbance attenuation.
For linear feedback control, Li et al. \cite{li2012GeneralizedExtended} proposed the generalized extended state observer-based control (GESOBC) for linear SISO/MIMO systems with mismatched disturbances, the mind behind which is similar to \cite{chen2011NonlinearDisturbance}, and the disturbance compensation gain also depends on the nominal control gain.
Regarding backstepping control, Sun et al. \cite{sun2014CompositeAdaptive,sun2014NonlinearDisturbance,sun2015GlobalOutput} combined the nonlinear disturbance observer with back-stepping technique to achieve mismatched disturbance attenuation, where estimated disturbances are integrated into the design of the backstepping controller in each step. 
As for SMC, Yang et al.~\cite{yang2013continuous,yang2013SlidingmodeControl} and Ginoya et al.~\cite{ginoya2014SlidingMode,ginoya2015DisturbanceObserver} have developed a series of disturbance observer-based SMCs, which achieve mismatched disturbance attenuation by designing a modified nonlinear sliding mode surface considering the estimated disturbance. However, the systems in SMC are limited to chains of integrators, and thus these methods are not suitable for
non-minimum phase systems.

These works have separately addressed mismatched disturbance attenuation, but have not offered a unified solution. Our previous work \cite{wang2023tieequilibrium} proposed a universal disturbance rejection scheme, equilibrium compensation-based control (ECBC), which is suitable for matched/mismatched disturbances and general linear systems, including non-minimum phase systems. In this article, we extend this method to nonlinear systems, resulting in a general disturbance rejection scheme named equilibrium adaptation-based control (EABC). 

Our core insight is that both the compensation methods for matched disturbances and those sophisticated methods for mismatched disturbances can be interpreted uniformly as the adaptation of equilibrium points to disturbances. Specifically, the EABC scheme involves three key steps: estimating the disturbance through existing disturbance observation techniques, calculating or optimizing the disturbed equilibrium points under the requirement of zero steady-state tracking error, and finally regulating the state to the disturbed equilibrium point using standard controllers.

\subsection{Contribution and Outline}
In summary, the main contributions of this work include:

1) We provide an interpretation for the mechanism of the track stand maneuver of STTW robots. To the best of the authors' knowledge, this work is the first study accomplishing the track stand maneuver with only one steering motor and one rear motor, without additional actuators.

2) We propose the EABC scheme for mismatched disturbance rejection of nonlinear systems, which can be seamlessly integrated with existing controllers and disturbance observers. In addition, we provide several conditions that ensure accurate tracking with bounded error or zero steady-state error.

3) We integrate the EABC with nonlinear MPC for track stand control of STTW robots. Experimental results demonstrate that our method significantly reduces tracking errors by several orders of magnitude.

The remainder of this article is organized as follows. Section~\ref{sec:EABC-framework} presents the problem formulation and the EABC disturbance rejection scheme. 
Section~\ref{sec:ts-model-UCEDT-validation} elucidates the underlying physical principle of the track stand maneuver and the implementation of EABC for the track stand control.
In Section~\ref{sec:experiment-validation}, the EABC scheme is validated through experiments with an STTW robot.
Finally, Section~\ref{sec:conclusion} concludes this article.

\subsection{Notations}
In this article, non-bold symbols denote scalars (e.g., $a$, $R$), bold lowercase symbols denote vectors (e.g., $\bm{x}$), and bold uppercase symbols represent matrices (e.g., $\bm{A}$). Especially, $\bm{I}_{n}$ represents the identity matrix of size $n$ and $\bm{0}_{n\times m}$ represents the zero matrix of size $n\times m$. The notation $|\bm{x}|$ represents element-wise absolute values, while the notation $\Vert \bm{x} \Vert$ represents the Euclidean norm. Additionally, the symbol $\preceq$ indicates the element-wise less than or equal to (e.g.,$\bm{x} \preceq \bm{y}, \bm{A} \preceq \bm{B}$). 

\section{The EABC Disturbance Rejection Scheme}\label{sec:EABC-framework}

\subsection{Problem Formulation}\label{subsec:problem-formualtion}
Consider a perturbed nonlinear system:
\begin{equation}\label{eq:nonlinear-disturbed-system}
    \begin{aligned}
    \dot{\bm{x}} &= \bm{f}(\bm{x}) + \bm{G}_u(\bm{x})\bm{u} + \bm{G}_{d}(\bm{x})\bm{d}\\
    \bm{y}_{m} &= \bm{h}_{m}(\bm{x})\\
    \bm{y}_{o} &= \bm{h}_{o}(\bm{x}),
    \end{aligned}
\end{equation}
where $\bm{x}\in \mathbb{R}^{n_x}$, $\bm{u}\in \mathbb{R}^{n_u}$, $\bm{d} \in \mathbb{R}^{n_d}$, $\bm{y}_m \in \mathbb{R}^{n_m}$, and $\bm{y}_o \in \mathbb{R}^{n_o}$ represent the state, input, disturbance, measurable output and controlled output, respectively. $\bm{f}(\bm{x}), \bm{G}_u(\bm{x}), \bm{G}_d(\bm{x}), \bm{h}_m(\bm{x}), \bm{h}_o(\bm{x})$ are smooth functions with appropriate dimensions.

\begin{definition}[\cite{chen1987robustness,chen2015disturbance,wang2023tieequilibrium}]\label{def:matched-disturbance}
A disturbance $d$ in $\bm{d}$ is classified as a matched disturbance if the corresponding column $\bm{g}_d(\bm{x})$ in $\bm{G}_d(\bm{x})$ lies within the range of $\bm{G}_u(\bm{x})$. That is, there exists a nonzero vector function $\bm{\lambda}(\bm{x})$ satisfying $\bm{G}_u(\bm{x}) \bm{\lambda}(\bm{x}) = \bm{g}_d(\bm{x})$. Otherwise, it is a mismatched disturbance.
\end{definition}

\begin{assumption}\label{ass:slow-varying-disturbance}
    The disturbance $\bm{d}$ is assumed to be a slow-varying disturbance. Namely, it is differentiable and bounded, satisfying $|\bm{d}(t)| \preceq \bm{d}_{\max}$, and its derivative $\dot{\bm{d}}$ is bounded, satisfying $|\dot{\bm{d}}(t)| \preceq \dot{\bm{d}}_{\max}$. 
\end{assumption}

\begin{remark}
    Assumption~\ref{ass:slow-varying-disturbance} is more general than those in previous studies on mismatched disturbance(\textit{e.g.} \cite{li2012GeneralizedExtended, wang2023tieequilibrium}), where the disturbance converges to a constant value as time goes to infinity. Representative disturbances include: constant parameter perturbation (\textit{e.g.} additional unknown payload), sensor bias (\textit{e.g.} inertial measurement unit equipment offset), and slow-varying external disturbances (\textit{e.g.} slow-varying ground inclination angle).
\end{remark}

\begin{figure}[t]
\centering
    \includegraphics[width=\columnwidth]{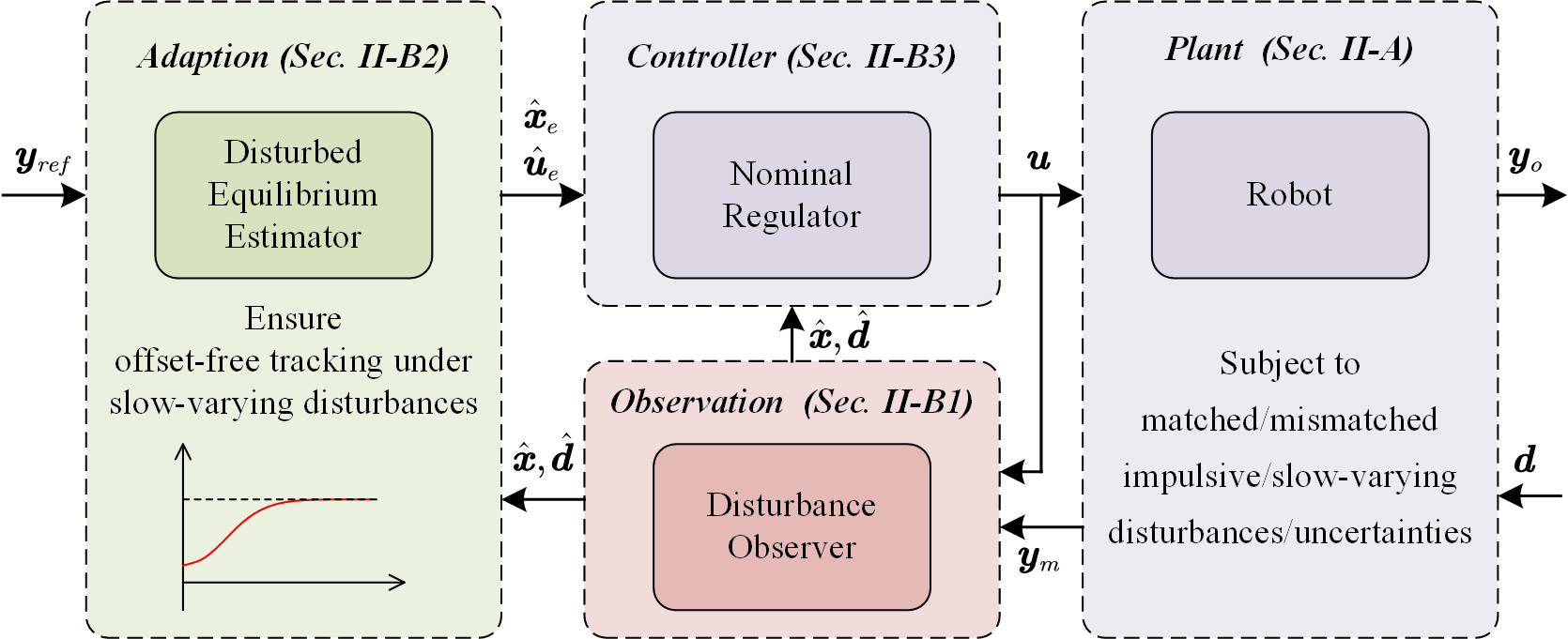}
    \caption{The general EABC disturbance rejection scheme.}
    \label{fig:poa-framework}
\end{figure}

\textbf{Problem statement:} For system \eqref{eq:nonlinear-disturbed-system} with slow-varying  disturbances that may be matched or mismatched, the objective is to design a disturbance rejection framework that achieves accurate tracking without steady-state error.

\textbf{Solution:} The EABC scheme is shown in Fig.~\ref{fig:poa-framework}, which is based on the idea that tracking accuracy only depends on the equilibrium point that the system converges to. Under slow-varying disturbances, the system first estimates the disturbances by general disturbance observers, then calculates an appropriate disturbed equilibrium trajectory ensuring tracking performance, and finally drives the state to it using a general regulation controller. 

\subsection{Components of the EABC Scheme}

\subsubsection{General Disturbance Observer} Define the extended state as $\bar{\bm{x}} = [\bm{x}^T, \bm{d}^T]^T$. The extended system is given by
\begin{equation}\label{eq:extended-system}
\begin{aligned}
    \dot{\bar{\bm{x}}} &= \bar{\bm{f}}(\bar{\bm{x}}) + \bar{\bm{G}}_u(\bar{\bm{x}})\bm{u} + \bar{\bm{B}}_d \dot{\bm{d}}\\
    \bm{y}_{m} &= \bar{\bm{h}}_{m}(\bar{\bm{x}}),
\end{aligned}
\end{equation}
with 
\begin{equation*}\begin{aligned}
    \bar{\bm{f}}(\bar{\bm{x}}) &= \begin{bmatrix}
        \bm{f}(\bm{x}) + \bm{G}_d(\bm{x})\bm{d} \\ \bm{0}_{n_d \times 1}
    \end{bmatrix},
    \bar{\bm{G}}_u(\bar{\bm{x}}) = \begin{bmatrix}
        \bm{G}_u(\bm{x}) \\ \bm{0}_{n_d \times n_u}
    \end{bmatrix},\\
    \bar{\bm{B}}_d &= \begin{bmatrix}
        \bm{0}_{n_x \times n_d} \\ \bm{I}_{n_d}
    \end{bmatrix}.
\end{aligned}\end{equation*}
We assume that the observer has the general form
\begin{equation}\label{eq:general-observer}
    \dot{\hat{\bar{\bm{x}}}} = \bm{\zeta}(\hat{\bar{\bm{x}}}, \bm{u}, \bm{y}_m),
\end{equation}
where $\hat{\bar{\bm{x}}} = [\hat{\bm{x}}^T, \hat{\bm{d}}^T]^T$ is the estimated state. Define the observation error as $\bm{e}_x = \bm{x} - \hat{\bm{x}}$, $\bm{e}_d = \bm{d} - \hat{\bm{d}}$, and $\bar{\bm{e}} = [\bm{e}_x^T, \bm{e}_d^T]^T$. Subtracting \eqref{eq:general-observer} from \eqref{eq:extended-system}, the error dynamics is denoted as
\begin{equation}\label{eq:general-observer-error}
    \dot{\bar{\bm{e}}} = \bm{\eta}(\bar{\bm{e}}, \dot{\bm{d}}),
\end{equation}
where $\bm{\eta}$ is a general function.
Typically, the disturbance observer is designed such that the error system \eqref{eq:general-observer-error} is stable. 
\begin{remark}
    This article does not focus on the disturbance observer but just employs it as a general module.
    Common disturbance observers include NDO~\cite{chen2000nonlinear, chen2004disturbance}, GESO~\cite{li2012GeneralizedExtended}, and ESO~\cite{han2009pid}. 
    If the disturbance model is also known, the extended state $\bar{\bm{x}}$ and the extended system \eqref{eq:extended-system} can be modified to include the disturbance model, such as \cite{guo2023CompositeDisturbance}. 
\end{remark}

\subsubsection{Disturbed Equilibrium Estimator}
Under the disturbance estimation $\hat{\bm{d}}$, to achieve an accurate steady-state tracking of the reference $\bm{y}_{ref}$, the estimation of disturbed equilibrium point, denoted by $(\hat{\bm{x}}_{e}, \hat{\bm{u}}_{e})$, should satisfy 
\begin{equation}\label{eq:equilibrium-output-equation}
    F_{\hat{\bm{d}}}(\hat{\bm{x}}_{e}, \hat{\bm{u}}_{e}) = 
    \left[
    \begin{aligned}
        &\bm{f}(\hat{\bm{x}}_{e}) + \bm{G}_u(\hat{\bm{x}}_{e})\hat{\bm{u}}_{e} + \bm{G}_{d}(\hat{\bm{x}}_e)\hat{\bm{d}} \\
        &\bm{h}_o(\hat{\bm{x}}_{e}) - \bm{y}_{ref}
    \end{aligned}
    \right]
    = \bm{0}.
\end{equation}
If \eqref{eq:equilibrium-output-equation} is ensured to have a unique solution, iteration methods can be used to obtain the root. Otherwise, if it has multiple solutions or no solution, the following optimization problem can be utilized to find an optimal solution:
\begin{equation}\label{eq:equilibrium-output-optimization}
    \underset{\hat{\bm{x}}_{e}, \hat{\bm{u}}_{e}}{\min} \  \left\Vert F_{\hat{\bm{d}}}(\hat{\bm{x}}_{e}, \hat{\bm{u}}_{e})\right\Vert_n + \left\Vert \hat{\bm{x}}_{e} \right\Vert_n + \left\Vert \hat{\bm{u}}_{e} \right\Vert_n,
\end{equation}
where $\Vert\cdot\Vert_n$ is a user-defined norm.

\subsubsection{General Regulation Controller}
The regulation controller in EABC has the form
\begin{equation}\label{eq:general-controller}
    \bm{u}_c(\hat{\bm{x}}) = \bm{\kappa}(\hat{\bm{x}} - \hat{\bm{x}}_{e}) + \hat{\bm{u}}_{e}.
\end{equation}
Generally, if the state $\bm{x}$ is known, the nominal controller $\bm{u}_c(\bm{x})$ is designed such that the equilibrium point $(\hat{\bm{x}}_{e}, \hat{\bm{u}}_{e})$ of the following nominal closed-loop system is stable:
\begin{equation}\label{eq:general-controller-closed}
\begin{aligned}
    \dot{\bm{x}} &= \bm{f}(\bm{x}) + \bm{G}_u(\bm{x})\bm{u}_c(\bm{x}) + \bm{G}_{d}(\bm{x})\hat{\bm{d}},\\
    \bm{y}_o &= \bm{h}_o (\bm{x}).
\end{aligned}
\end{equation}
\begin{remark}
The type of regulation controller \eqref{eq:general-controller} is not specified in the EABC, and some typical controllers can be used, such as model predictive control (MPC)~\cite{morari2012nonlinear}, feedback linearization control~\cite{khalil2015nonlinear}. In our implementation for track stand control, we choose a nonlinear MPC as the regulation controller, as introduced in Section~\ref{subsec:nmpc}.
\end{remark}

So far, all the components of EABC have been presented. To introduce the properties of EABC, we define the nominal regulation error system in advance. 
Let $\tilde{{\bm{x}}} = \bm{x} - \hat{\bm{x}}_e$ and $\tilde{\bm{y}}_o = \bm{y}_o - \bm{y}_{ref}$ be the regulation error and the tracking error.
Subtracting \eqref{eq:equilibrium-output-equation} from \eqref{eq:general-controller-closed} gives the nominal error system:
\begin{equation}\label{eq:general-error-system}
\begin{alignedat}{2}
    \dot{\tilde{{\bm{x}}}} 
    &= \bm{A}\tilde{{\bm{x}}} + \bm{G}_u(\bm{x}) \bm{\kappa}(\tilde{{\bm{x}}}),\\
    {\tilde{{\bm{y}}}_o} &= \bm{C}_o\tilde{{\bm{x}}},
\end{alignedat}
\end{equation}
with
\begin{equation*}
    \bm{A} = \left.\left(\frac{\partial \bm{f}}{\partial \bm{x}}
    + \frac{\partial \bm{G}_u}{\partial \bm{x}}{\hat{\bm{u}}_e} 
    + \frac{\partial \bm{G}_d}{\partial \bm{x}}{\hat{\bm{d}}} \right)\right|_{\bm{x} 
    = \hat{\bm{x}}_e},\quad
    \bm{C}_o = \left.\frac{\partial\bm{h}_o}{\partial \bm{x}}\right|_{\bm{x} = \hat{\bm{x}}_e}.
\end{equation*}
Additionally, we define the nominal perturbed regulation error system as
\begin{equation}\label{eq:general-perturbed-error-system}
    \dot{\tilde{{\bm{x}}}} = \bm{A}\tilde{{\bm{x}}} + \bm{G}_u(\bm{x}) \bm{\kappa}(\tilde{{\bm{x}}}) + \bm{\varepsilon},
\end{equation}
where $\bm{\varepsilon}$ denotes a general state perturbation.

\subsection{Properties of EABC}

\begin{theorem}\label{th:EABC_bounded_zeroerror}
    Consider the system~\eqref{eq:nonlinear-disturbed-system} controlled by the EABC \eqref{eq:general-observer}--\eqref{eq:equilibrium-output-equation}--\eqref{eq:general-controller}. If Assumption~\ref{ass:slow-varying-disturbance} holds and the EABC satisfies the following conditions: \begin{enumerate}
        \renewcommand{\labelenumi}{\arabic{enumi})}
        \item The observation error system \eqref{eq:general-observer-error} is input-to-state stable;
        \item The equilibrium equation \eqref{eq:equilibrium-output-equation} has more than one solution;
        \item The nominal regulation error system \eqref{eq:general-perturbed-error-system} is input-to-state stable;
        \item Function $\bm{\kappa}(\tilde{\bm{x}})$ is Lipschitz continuous;
        \item Function $\bm{G}_u(\bm{x})$ and $\bm{G}_d(\bm{x})$ are bounded, denoted as $\left\vert\bm{G}_u(\bm{x})\right\vert \preceq \bm{G}_{u,b}, \left\vert\bm{G}_d(\bm{x})\right\vert \preceq \bm{G}_{d,b}$;
    \end{enumerate}
    then the following conclusions hold: 
    \begin{enumerate}
        \renewcommand{\labelenumi}{\alph{enumi})}
        \item The controlled output $\bm{y}_o$ tracks the reference $\bm{y}_{ref}$ with bounded error.
        \item Moreover, if the disturbance derivative $\dot{\bm{d}}$ converges to zero, then the tracking error $\tilde{\bm{y}}_o$ will also converge to zero.
    \end{enumerate}
\end{theorem}
\begin{IEEEproof}
    Since $\bm{\kappa}(\hat{\bm{x}} - \hat{\bm{x}}_e) = \bm{\kappa}(\bm{x}-\bm{e}_x - \hat{\bm{x}}_e) = \bm{\kappa}(\tilde{\bm{x}} - \bm{e}_x)$, by substituting $\bm{u}_c(\hat{\bm{x}})$ for $\bm{u}$ in \eqref{eq:nonlinear-disturbed-system}, the closed-loop system under EABC \eqref{eq:general-observer}--\eqref{eq:equilibrium-output-equation}--\eqref{eq:general-controller} is given by
    \begin{equation}\begin{aligned}\label{eq:EABC-closedloop-system}
        \dot{\bm{x}} &= \bm{f}(\bm{x}) + \bm{G}_u(\bm{x})\bm{u}_c({\bm{x}}) + \bm{G}_d(\bm{x}) \hat{\bm{d}} + \bm{\epsilon}, \\
        \bm{\epsilon} &= \bm{G}_d(\bm{x}) \bm{e}_d +\bm{G}_u(\bm{x})(\bm{\kappa}(\tilde{\bm{x}} - \bm{e}_x) - \bm{\kappa}(\tilde{\bm{x}})).
    \end{aligned}\end{equation}
    Subtracting \eqref{eq:equilibrium-output-equation} from \eqref{eq:EABC-closedloop-system} gives
    \begin{equation} \begin{aligned}
        \dot{\tilde{\bm{x}}} &= \bm{A}\tilde{\bm{x}} + \bm{G}_u(\bm{x}) \bm{\kappa}(\tilde{{\bm{x}}}) + \bm{\epsilon},\\
        {\tilde{\bm{y}}_o} &= \bm{C}_o\tilde{\bm{x}}.
    \end{aligned} \end{equation}
    According to Assumption~\ref{ass:slow-varying-disturbance} and Condition 1), the estimation errors ${\bm{e}}_x,\bm{e}_d$ are bounded, denoted as $|{\bm{e}}_x| \preceq {\bm{e}}_{x,b}, |{\bm{e}}_d| \preceq {\bm{e}}_{d,b}$. According to Condition 4), for the $i\text{-th}$ row of $\bm{\kappa}$, there exists a positive constant ${m}_i > 0$ such that 
    \begin{equation}
        \left\vert {\kappa}_i(\tilde{\bm{x}} - \bm{e}_x) - {\kappa}_i(\tilde{\bm{x}}) \right\vert 
        \le {m}_i \left\Vert \bm{e}_{x} \right\Vert 
        \le {m}_i \left\Vert \bm{e}_{x,b} \right\Vert.
    \end{equation}
    Thus, we have
    \begin{equation}
        \left\vert \bm{\kappa}(\tilde{\bm{x}} - \bm{e}_x) - \bm{\kappa}_i(\tilde{\bm{x}}) \right\vert 
        \preceq \bm{m} \left\Vert \bm{e}_{x,b} \right\Vert,
    \end{equation}
    where $[\bm{m}]_i = m_i$.
    According to Condition 5), $\bm{G}_u(\bm{x})$ and $\bm{G}_d(\bm{x})$ are bounded, so $\bm{\epsilon}$ is bounded and satisfies
    \begin{equation}\begin{aligned}\label{eq:epsilon_bound}
        |\bm{\epsilon}| &= \left\vert\bm{G}_d(\bm{x}) \bm{e}_d + \bm{G}_u(\bm{x})(\bm{\kappa}(\tilde{\bm{x}} - \bm{e}_x) - \bm{\kappa}(\tilde{\bm{x}}))\right\vert \\
        &\preceq \bm{G}_{d,b} \bm{e}_{d,b} + \bm{G}_{u,b} \bm{m} \left\Vert \bm{e}_{x,b} \right\Vert.
    \end{aligned}\end{equation}
    According to Condition 3), the regulation error state $\tilde{\bm{x}}$ and the tracking error $\tilde{\bm{y}}_o$ are also bounded. Thus Conclusion a) holds.

    Furthermore, if $\dot{\bm{d}}\to \bm{0}$ as $t \to \infty$, since  \eqref{eq:general-observer-error} is input-to-state stable, we have $\bar{\bm{e}} \to \bm{0}$. From \eqref{eq:epsilon_bound}, $\bm{\epsilon} \to \bm{0}$. Applying the property of input-to-state stability of \eqref{eq:general-perturbed-error-system}, we can deduce that $\tilde{\bm{x}} \to \bm{0}, \tilde{\bm{y}}_o \to \bm{0}$, which gives Conclusion b).
\end{IEEEproof}

\begin{remark}
    According to \eqref{eq:epsilon_bound}, even if the disturbance derivative $\dot{\bm{d}}$ does not converge to zero, we can still decrease the tracking error by improving the performance of the disturbance observer, specifically by reducing the estimation error $\bar{\bm{e}}$ closer to zero.
\end{remark}

\begin{remark}
    Since \eqref{eq:equilibrium-output-equation} is a necessary condition for equilibrium points and accurate tracking, if the equilibrium equation \eqref{eq:equilibrium-output-equation} does not have a solution, the system cannot accurately regulate the controlled output $\bm{y}_o$ to the reference $\bm{y}_{ref}$.
\end{remark}

\subsection{Relation with Existing Methods}
To the best knowledge of the authors, although the mind behind EABC is intuitive, it has not been reported in existing literature, except for our prior study of ECBC~\cite{wang2023tieequilibrium,wang2023iciraequilibrium}, which is the linear case of EABC. EABC draws inspiration from nonlinear disturbance observer-based control (NDOBC)~\cite{chen2004disturbance, chen2015disturbance,guo2005DisturbanceAttenuation,chen2011NonlinearDisturbance} but provides a novel point of view.

\subsubsection{Relation with ECBC} EABC and ECBC~\cite{wang2023tieequilibrium,wang2023iciraequilibrium} are based on the same process: first estimating the disturbance by disturbance observers, then solving the disturbed equilibrium point and finally regulating the state to that equilibrium point by general controllers. However, ECBC is only suitable for linear systems, while EABC is applicable to nonlinear systems.

\subsubsection{Relation with NDOBC} From the original concept of NDOBC~\cite{chen2004disturbance}, EABC can be considered a special case of NDOBC. From the perspective of specific applications, EABC differs fundamentally in its underlying principles compared to NDOBCs for matched disturbances~\cite{chen2004disturbance,guo2005DisturbanceAttenuation} and mismatched disturbances~\cite{chen2011NonlinearDisturbance}. These NDOBCs try to design a controller with respect to the state and the estimated disturbance, with the formulation:
\begin{align}
    \bm{u}_{\text{NDOBC}} = \bm{\kappa}(\hat{\bm{x}}) - \bm{\beta}(\hat{\bm{x}})\hat{\bm{d}},
\end{align}
where $\bm{\kappa}(\hat{\bm{x}})$ is the feedback control law without considering disturbances and $\bm{\beta}(\hat{\bm{x}})$ is the disturbance compensation gain to be designed. For matched disturbances, $\bm{\beta}(\hat{\bm{x}})$ is only related to $\bm{G}_u(\bm{x})$ and ${\bm{G}_d(\bm{x})}$, while for mismatched disturbances, $\bm{\beta}(\hat{\bm{x}})$ has a complex form coupled with $\bm{\kappa}(\hat{\bm{x}})$.

In contrast, EABC from the view on \textit{disturbed equilibrium trajectory} (time-varying equilibrium point) with the formulation:
\begin{equation}
    \bm{u}_{\rm{EABC}} = \bm{\kappa}(\hat{\bm{x}} - \hat{\bm{x}}_e) + \hat{\bm{u}}_e, \text{with $(\hat{\bm{x}}_e,\hat{\bm{u}}_e)$ satisfying \eqref{eq:equilibrium-output-equation}},
\end{equation}
which is suitable for both matched and mismatched disturbances. Furthermore, whether disturbances can be eliminated from the steady state of controlled output depends on the solvability of \eqref{eq:equilibrium-output-equation}, and is independent of $\bm{\kappa}(\hat{\bm{x}}, \hat{\bm{x}}_e)$.

\section{Track Stand Model and EABC Design}\label{sec:ts-model-UCEDT-validation}
In Section~\ref{subsec:TS-model}, we first establish the dynamics of the STTW robot using the Newton-Euler method, by which we can obtain an intuitive physical interpretation of the mechanism of track stand maneuver. 
Then, we detail the EABC design for track stand control in Section~\ref{subsec:EABC-for-track stand}.

\subsection{Track Stand Model}\label{subsec:TS-model}

\begin{figure*}[t]
  \centering
  \includegraphics[width = 0.95\textwidth]{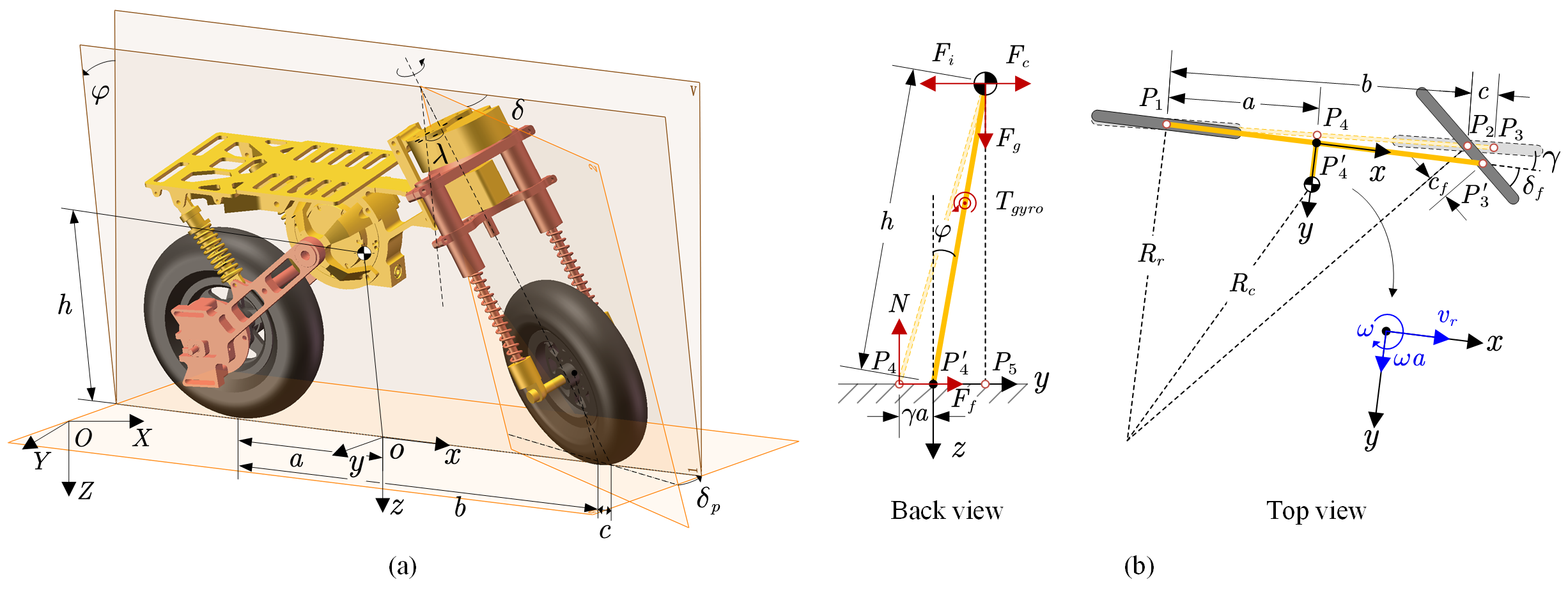}
  \caption{(a) Schematic of the STTW robot. (b) Back view and top view of the STTW robot.}
  \label{fig:model_schematic_tbview}
\end{figure*}

\begin{table}[thb]
\caption{Notations for the STTW robot\label{tab:notations}}
\centering
\begin{tabular}{| c | c |}
    \hline
    Symbols    & Descriptions\\
    \hline
    $\varphi$  & roll angle\\
    \hline
    $\delta$   & steering angle\\
    \hline
    $\delta_p$ & projection of the steering angle on the horizontal plane\\
    \hline
    $\lambda$  & caster angle\\
    \hline
    $\gamma$   & yaw deviation angle caused by steering rotation\\
    \hline
    $P_1/P_2$  & contact points of the \par rear/front wheel and the ground\\
    \hline
    $P_3$      & intersection point of the \par steering axis with the ground\\
    \hline
    $P_4$      & projection point of the COM \par onto the wheelbase line\\
    \hline
    $h$        & height of the COM\\
    \hline
    $a$        & distance from COM to the contact point of the rear wheel\\
    \hline
    $b$        & wheelbase\\
    \hline
    $c$        & trail length corresponding to $\delta = 0$\\
    \hline
    $c_f$      & trail length\\
    \hline
    $r$        & radius of the wheels\\
    \hline
    $R_r$      & turning radius of the rear wheel\\
    \hline
    $m$        & total mass\\
    \hline
    $I_b$      & total moment of inertia about the roll axis\\
    \hline
    $\omega$   & yaw angular velocity\\
    \hline
    $v_r$      & rear wheel velocity\\
    \hline
    $a_{o}$    & translational acceleration of frame $o\text{-}xyz$\\
    \hline
    $\tau_r$   & torque of the rear wheel\\
    \hline
    $I_r$      & inertia of the rear wheel\\
    \hline
    $s$        & distance traveled by the robot\\
    \hline
\end{tabular}
\end{table}

The schematic of the STTW robot developed in this study is shown in Fig.~\ref{fig:model_schematic_tbview}a. The notation used is listed in Table~\ref{tab:notations}. We make the following assumptions during modeling: (i) the front and rear suspensions are locked; (ii) there is no lateral slippage between the wheels and the ground; (iii) the thicknesses of the wheels are negligible. 
Detailed steps of the modeling process are provided in the Appendix\ref{sec:appendix-model}. Here, we focus on presenting the key steps and results. 

According to Fig.~\ref{fig:model_schematic_tbview}b, consider the roll motion, the robot is subject to gravity $F_g$, inertial force $F_i$, centrifugal force $F_c$ and normal force $N$. The roll dynamics is given by
\begin{equation}\label{eq:main-euler}
    (I_t+mh^2) \ddot{\varphi} = F_g h\sin \varphi + F_c h \cos \varphi + N \gamma a - F_i h \cos \varphi,
\end{equation}
where $F_i = m a_{o,y}$ and $a_{o,y}$ includes a term $\frac{a \cos \lambda}{b \cos \varphi} \dot{v}_r \tan \delta$ produced by the longitudinal acceleration, as shown in \eqref{eq:a_oy} of Appendix\ref{sec:appendix-model}. 
By selecting the state as $\bm{x} = [s, v_r, \delta, \varphi, \dot{\varphi}]^T$, the control input as $\bm{u} = [\tau_r, \dot{\delta}]^T$, and the disturbance as $\bm{d} = [d_{r1} r + d_{r2}, d_\varphi]^T$, the dynamics of the system can be represented as $\dot{\bm{x}} = \bm{f}_{sys}(\bm{x},\bm{u},\bm{d})$, with $\bm{f}_{sys}$ given by \eqref{eq:f_sys} in Appendix\ref{sec:appendix-model}.
It can be validated that around a given track stand equilibrium point $\bm{x}_e = [s_e, 0, \delta_e, \varphi_e, 0]^T, \bm{u}_e = [0,0]^T, \delta_e \ne 0$, satisfying $\bm{f}_{sys}(\bm{x}_e, \bm{u}_e, \bm{0}) = \bm{0}$, the locally linearized system is controllable.

\textit{Interpretation of the mechanism behind the track stand}: Through \eqref{eq:main-euler}, the physical principle governing the track stand maneuver can be stated as follows: The inertial force generated by the forward and backward movement of the rear wheel, under the geometric constraints introduced by steering, produces a lateral force component $m a_{o,y}$, which is leveraged to counteract the gravitational force, thereby enabling maintaining balance at stand still.
Furthermore, with a natural steering angle, $\delta \in (-\pi, \pi)$, the larger the steering angle, the more ``balance torque" the track stand maneuver can provide. If the steering angle is equal to zero, no ``balance torque" can be generated.

\subsection{EABC Design for Track Stand Control}\label{subsec:EABC-for-track stand}

\begin{figure}[t]
\centering
    \includegraphics[width=\columnwidth]{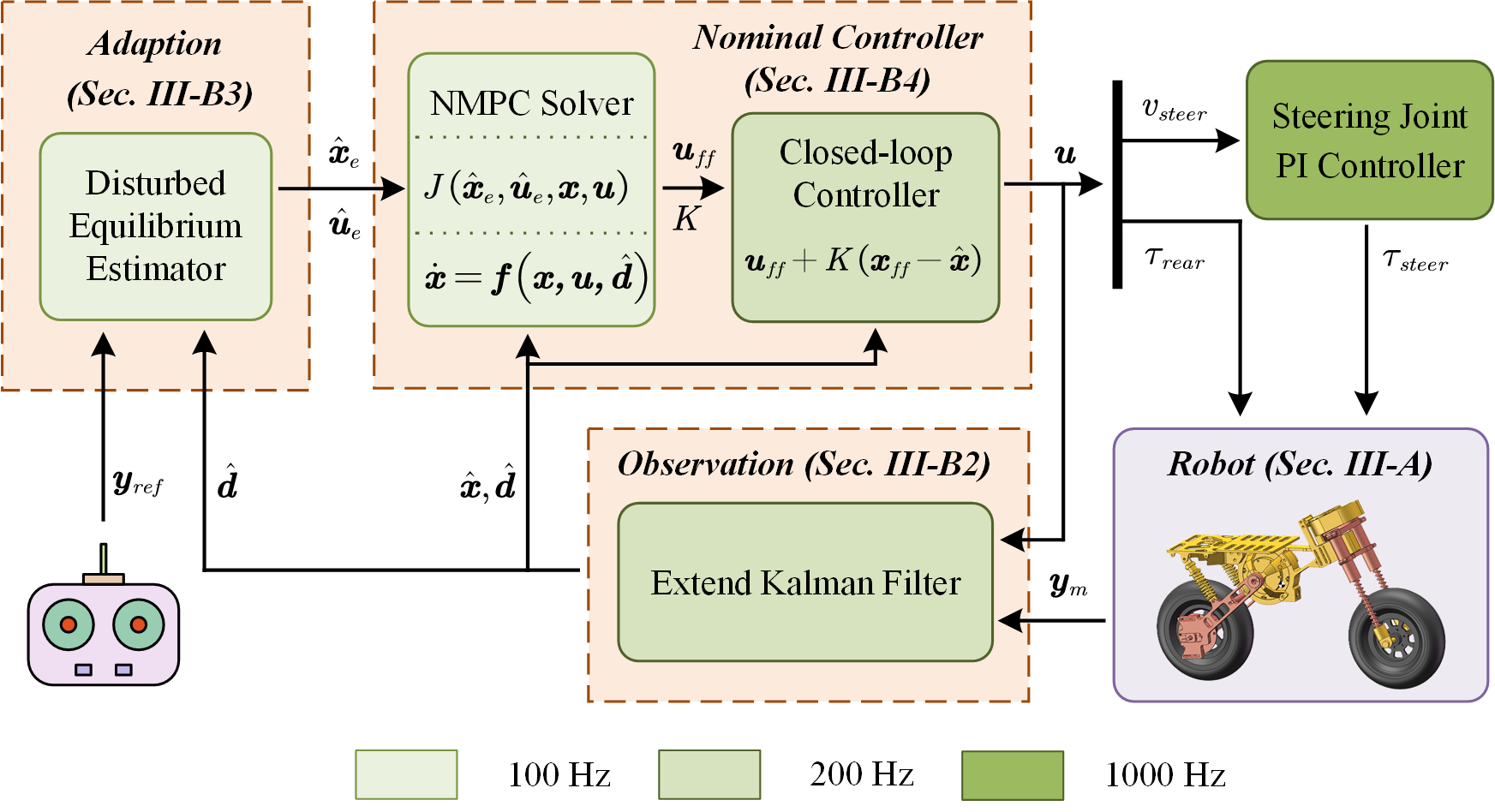}
    \caption{Block diagram of the control system for the robot.}
    \label{fig:block-diagram-STTW}
\end{figure}

The detailed EABC design for the STTW robot is illustrated in Fig.~\ref{fig:block-diagram-STTW}, comprising an extended Kalman filter for disturbance observation, a disturbed equilibrium estimator, and a nonlinear MPC as the nominal regulator.

\subsubsection{System Specification}
For the track stand task, the robot is expected to track a traveled distance $s_{ref}$ and a steering angle $\delta_{ref}$. According to dynamics \eqref{eq:f_sys}, the terms in the perturbed nonlinear system \eqref{eq:nonlinear-disturbed-system} are given by
\begin{align}
    &\bm{f}(\bm{x}) = 
    \begin{bmatrix}
    	x_2   \\
    	0     \\
    	0     \\
    	x_5\\
    	\beta_1(\bm{x})
    \end{bmatrix},
    \bm{G}_u(\bm{x}) = 
        \begin{bmatrix}
        	0                     & 0\\
        	\beta_2               & 0\\
        	0                     & 1\\
        	0                     & 0\\
            \beta_3(\bm{x})       & \beta_4(\bm{x})
        \end{bmatrix}, \label{eq:fx_Gux}\\
    &{\bm{G}}_{d}(\bm{x}) = \begin{bmatrix}
        0 & \beta_2 & 0 & 0 & \beta_3(\bm{x})\\
        0 & 0 & 0 & 0 & \beta_5
    \end{bmatrix}^T,\label{eq:Gdx}\\
    &\bm{h}_m(\bm{x}) = \bm{C}_m\bm{x},
    \bm{h}_o(\bm{x}) = \bm{C}_o\bm{x},\\
    &\bm{C}_m = \begin{bmatrix}
        \bm{I}_{4}, \bm{0}_{4\times1}
    \end{bmatrix},
    \bm{C}_o = \begin{bmatrix}
        1\ 0\ 0\ 0\ 0\\
        0\ 0\ 1\ 0\ 0\\
    \end{bmatrix},
\end{align}
where $\beta_1$, $\cdots$, $\beta_5$ are the corresponding constants or functions, provided in the appendix. It can be validated that both $\bm{G}_u$ and $\bm{G}_d$ are bounded.

\subsubsection{Disturbance Observer} 
For the extended system \eqref{eq:extended-system}, the nonlinear extended state observer is designed as follows:
\begin{equation}\label{eq:disturbance-observer}
\begin{aligned}
    &\dot{\hat{\bar{\bm{x}}}} = \bar{\bm{f}}(\hat{\bar{\bm{x}}}) + \bar{\bm{G}}_u(\hat{\bar{\bm{x}}})\bm{u} + \bm{L}(\bar{\bm{x}}, \bm{u})\bar{\bm{C}}_m(\hat{\bar{\bm{x}}} - \bar{\bm{x}}),\\
    &\bar{\bm{C}}_m = \begin{bmatrix}
        \bm{I}_{5\times5} & \bm{0}_{5\times 1} \\ \bm{0}_{1\times 5} & 0
    \end{bmatrix}.
\end{aligned}
\end{equation}
Then the observation error $\bar{\bm{e}} = \hat{\bar{\bm{x}}} - \bar{\bm{x}}$ satisfies
\begin{equation}\label{eq:design-estimation-error-system}
    \dot{\bar{\bm{e}}} = [\bm{A}(\bar{\bm{x}}, \bm{u}) + \bm{L}(\bar{\bm{x}}, \bm{u})\bar{\bm{C}}_m]\bar{\bm{e}} + \bar{\bm{B}}_d \dot{\bm{d}},
\end{equation}
where $\bm{A}(\bar{\bm{x}},\bm{u})$ is the Jacobin matrix of $\bar{\bm{f}} + \bar{\bm{G}}_u\bm{u}$ with respect to $\bar{\bm{x}}$ and $ \bm{L}(\bar{\bm{x}}, \bm{u})$ is the observation gain, which is appropriately designed to satisfy that $\bm{A}(\bar{\bm{x}},\bm{u}) +  \bm{L}(\bar{\bm{x}}, \bm{u})\bar{\bm{C}}_m$ is Hurwitz. $\bm{L}(\bar{\bm{x}},\bm{u})$ can be calculated by pole placement (Luenberger observer) or by Riccati recursion (steady-state extended Kalman filter). We use the second method in our experiments. 

According to Lemma 4.5 in \cite{khalil2015nonlinear}, the linear estimation error system \eqref{eq:design-estimation-error-system} is globally exponentially stable at the origin, and thus the system is input-to-state stable, satisfying Condition 1) in Theorem~\ref{th:EABC_bounded_zeroerror}.

\subsubsection{Disturbed Equilibrium Estimator} 
For track stand control, to estimate disturbed equilibrium, the equilibrium equation \eqref{eq:equilibrium-output-equation} should be solved. 
According to Definition~\ref{def:matched-disturbance}, $d_{r}$ is a matched disturbance, while $d_{\varphi}$ is a mismatched disturbance. 
Under the estimated disturbance $\hat{\bm{d}} = [\hat{d}_{r}, \hat{d}_{\varphi}]^T$, the solution to the equilibrium equation \eqref{eq:equilibrium-output-equation} has the form of
\begin{align}
    \hat{\bm{x}}_{e} &= [s_{ref}, 0, \delta_{ref}, \hat{\varphi}_{e}, 0]^T,\ 
    \hat{\bm{u}}_{e} = [-\hat{d}_{r}, 0]^T, \label{eq:hat_xe_ue}\\
    \hat{\varphi}_{e} &= \sin^{-1}\left(\frac{ mg\frac{ac}{b}c_\lambda \delta - \hat{d}_\varphi}{mgh} \right)\label{eq:hat_phi_e}.
\end{align}
Condition 5) is satisfied, provided that \eqref{eq:hat_xe_ue} and \eqref{eq:hat_phi_e} are valid.

\subsubsection{GNMS-Based Nonlinear MPC}\label{subsec:nmpc}
To make use of the system's nonlinear dynamics, we choose nonlinear MPC as the regulation controller, which has been validated to be effective in robots such as quadruped robots and drones. The procedure of nonlinear MPC is iteratively solving a finite-horizon, receding nonlinear optimal control problem (NOCP), which can be solved by shooting methods or direct collocation, where the original NOCP is transcribed into a nonlinear programming problem solved by a general solver. However, such methods are not fast enough for real-time MPC applications. Recently, GNMS methods \cite{neunert2016fast,giftthaler2018family,neunert2018whole} have been developed to solve NOCPs by leveraging their sparsity structure using Riccati recursion, ensuring that their computing speed meets the requirement of running in real time.

The receding NOCP problem is given by
\begin{equation}\label{eq:MPC-NOCP}
\begin{aligned}
    & \underset{\bm{x},\bm{u}}{\text{min}}  &&J(\bm{x},\bm{u}) = \int_{0}^{t_f} \tilde{\bm{x}}(t)^T Q \tilde{\bm{x}}(t) + \tilde{\bm{u}}(t)^T R \tilde{\bm{u}}(t) {\rm{d}}t \\
    &&& \ \ \ \ \ \ \ \ \ \ \ \  + \tilde{\bm{x}}(t_f)^T H \tilde{\bm{x}}(t_f)  \\
    & \text{s.t.}  &&\dot{\bm{x}} = \bm{f}(\bm{x}) + \bm{G}_u(\bm{x})\bm{u}+ \bm{G}_{d}(\bm{x})\hat{\bm{d}}\\
    &&& \bm{x}(0) = \hat{\bm{x}},
\end{aligned}
\end{equation}
where $\tilde{\bm{x}} = \bm{x} - \hat{\bm{x}}_{e}$, $\tilde{\bm{u}} = \bm{u} - \hat{\bm{u}}_{e}$, 
$\bm{Q}$ is a positive semidefinite matrix and $\bm{R}$ is a positive definite weighting matrix. In this context, we assume that the conditions about the cost function for the input-to-state stability of MPC, as outlined in~\cite{marruedo2002InputstateStable,limon2009input}, are satisfied. The goal of the NOCP is to find the following feedforward and feedback control law to stabilize the closed-loop system:
\begin{equation} \label{eq:MPC-controller}
    \bm{u}(t) = \bm{u}_{ff}(t) + \bm{K}(t) (\bm{x} - \bm{x}_{ref}),
\end{equation}
where $\bm{u}_{ff}(t)$ is the feedforward control and $
\bm{K}(t)$ is the feedback control gain.

\section{Experimental Validation of the EABC Scheme}\label{sec:experiment-validation}

\begin{figure}[t]
    \centering
    \includegraphics[width=\columnwidth]{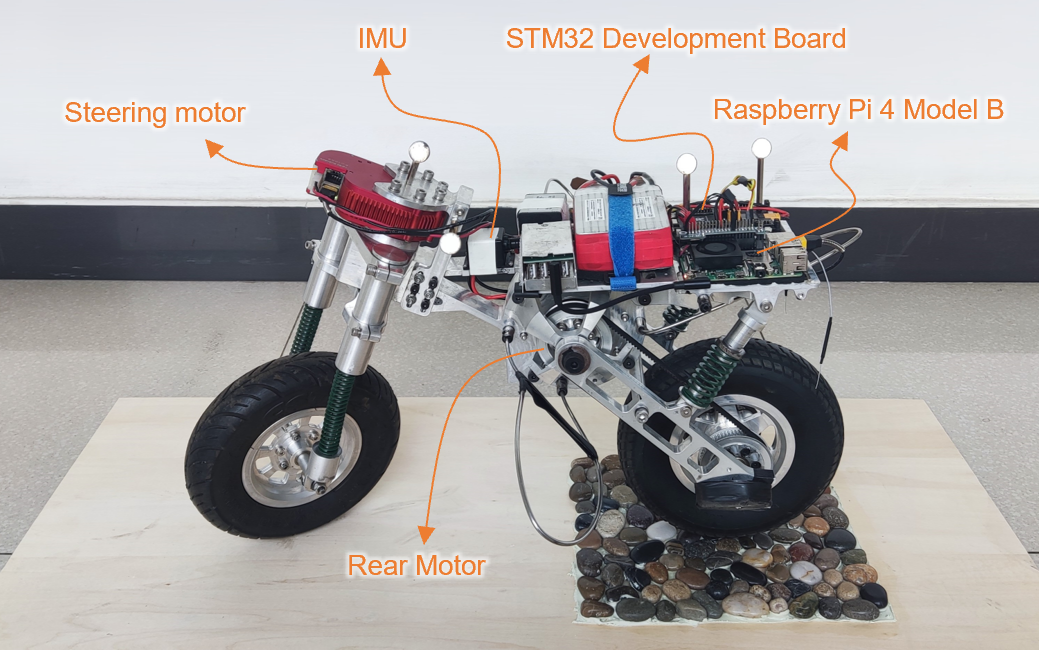}
    \caption{The STTW robot test bench.}
    \label{fig:exp_robot_photo}
\end{figure}

\begin{table}
    \caption{Physical parameter values of the STTW robot.}
    \label{tab:physical-params}
    \begin{center}
    \begin{tabular}{|c|c||c|c||c|c|}
    \hline
    Parameter & Value & Parameter & Value & Parameter & Value\\
    \hline
    $a$ (m) & 0.140 & $h$ (m) & 0.2 & $m$ (kg) & 7.4\\
    \hline
    $b$ (m) & 0.408 & $r$ (m) & 0.1   & $I_t$ ($\mathrm{kg\cdotp m^2}$) & 0.356\\
    \hline
    $c$ (m) & 0.024 & $\lambda\ (^\circ)$ & 25 & $I_r$ ($\mathrm{kg\cdotp m^2}$) & 0.02\\
    \hline
    \end{tabular}
    \end{center}
\end{table}

\begin{figure*}[!t]    
    \centering
    \includegraphics[width=\textwidth]{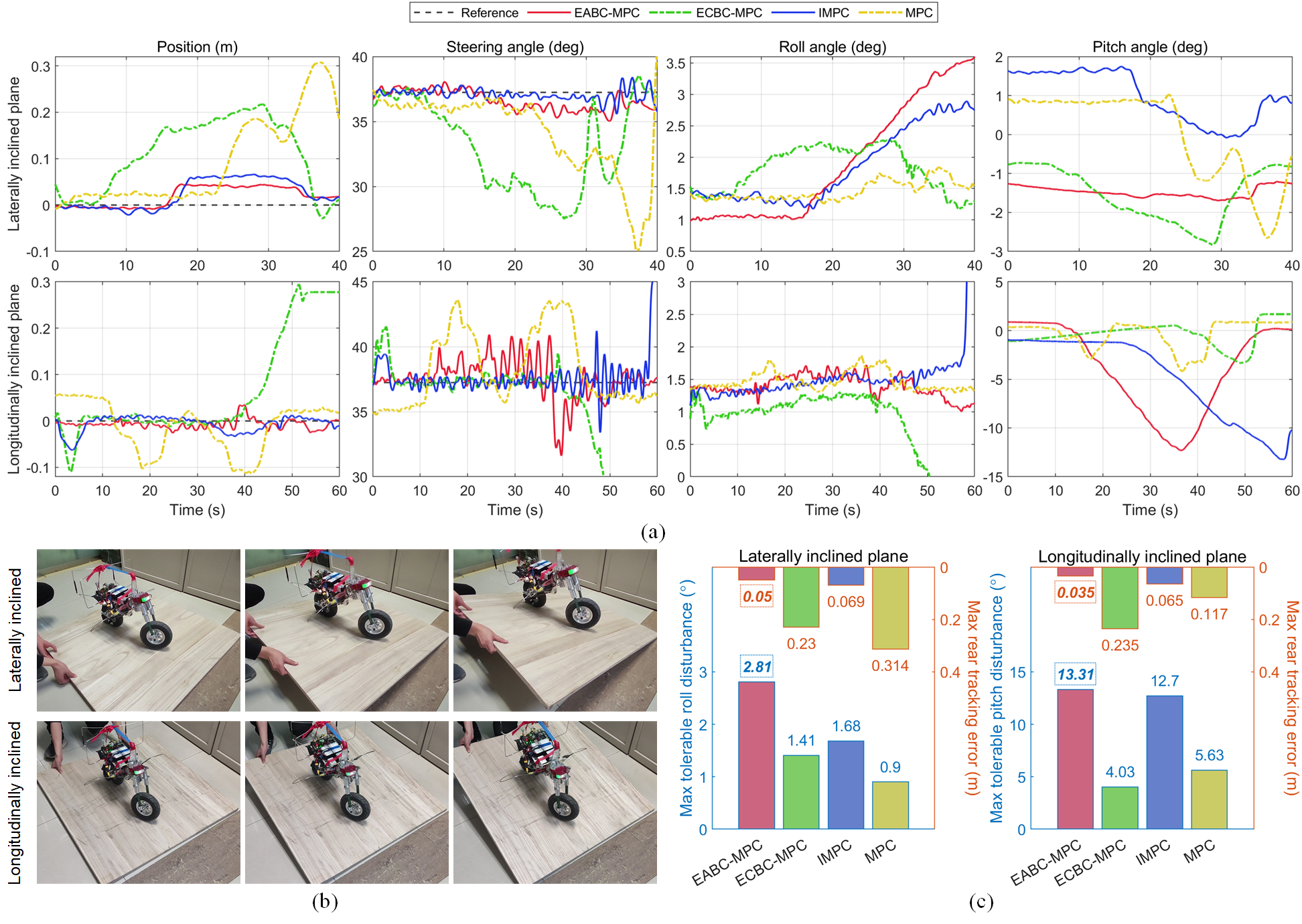}
    \caption{(a) Experimental results of track stand on a laterally/longitudinally inclined plane under four controllers. For clarity of illustration, all signals are filtered by a Butterworth low-pass filter with a cutoff frequency of $\text{0.5 Hz}$ using the \textit{filtfilt} function of MATLAB. (b) Snapshots of the experiments. (c) Maximum tolerable roll/pitch disturbances and maximum rear position tracking error during experiments with four controllers. The best values are in bold and surrounded by boxes.}
    \label{fig:exp-inclinedplane-entire}
\end{figure*}

\begin{figure*}[!t]
    \centering
    \includegraphics[width=\textwidth]{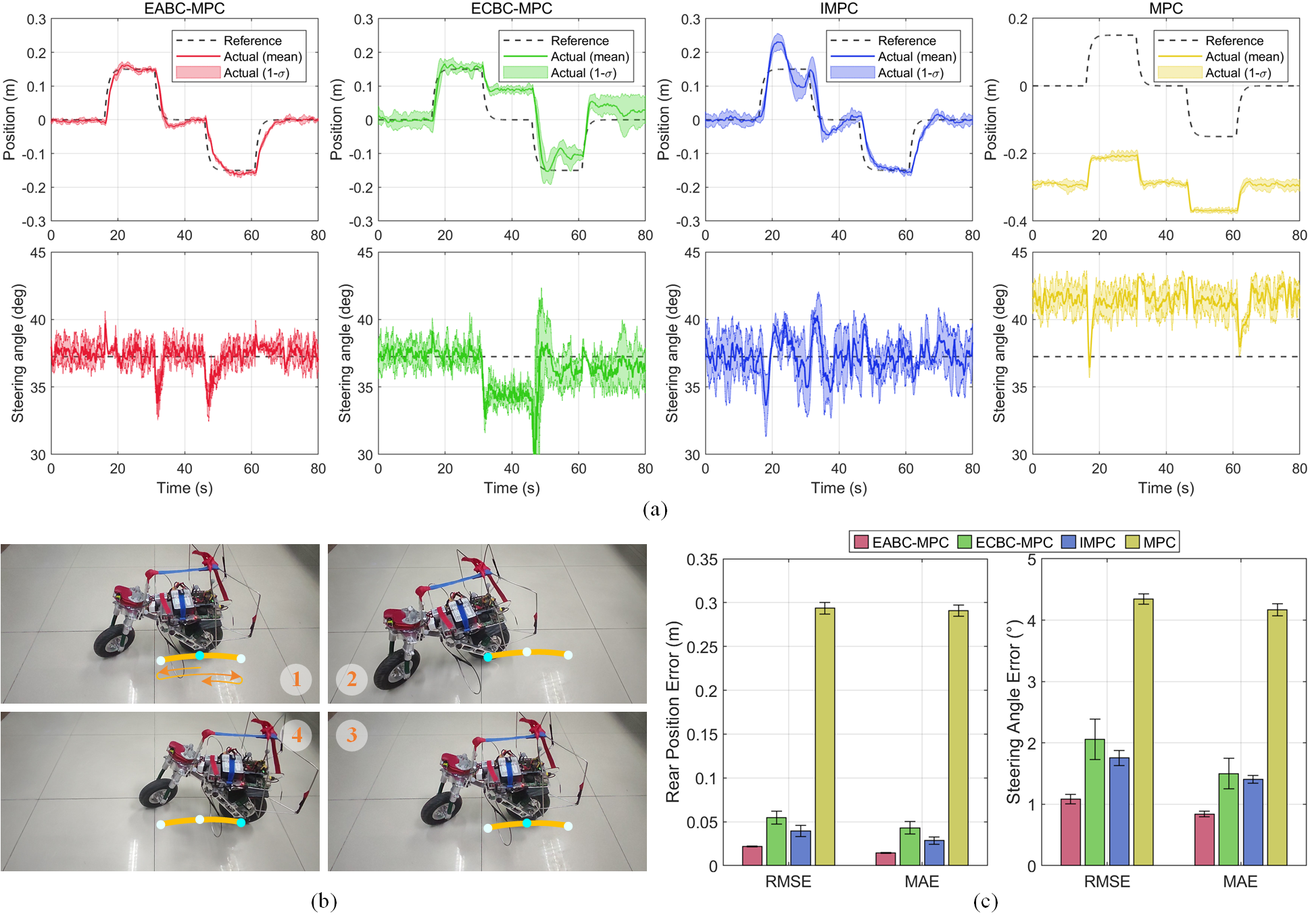}
    \caption{(a) Experimental results of the rear position and steering angle tracking task. Experiments for each controller are conducted five times. The mean values and $1\text{-}\sigma$ regions of the five sets of experimental data are plotted by solid lines and shaded areas in light colors. (b) Snapshots of the experiments. (c) RMSE and RAE of the rear position tracking error and the steering angle tracking error under four methods. All RMSEs and RAEs are calculated over five experimental runs, and the error bars indicate the standard deviation of the five runs.}
    \label{fig:exp-postrack-entire}
\end{figure*}

In this section, we deploy the EABC on an STTW robot test bench, as depicted in Fig.~\ref{fig:exp_robot_photo}, to validate its disturbance rejection capability through two experiments. 
The first scenario is the \textit{inclined plane track stand} (Section~\ref{subsec:adaptation_exp_inclined_plane}), which validates the capability to reject a varying external disturbance. The second scenario is the \textit{rear position tracking} (Section~\ref{subsec:adaptation-exp-rear-tracking}), where the robot tracks a varying reference. 

\subsection{Experimental Setup}

\subsubsection{Hardware Description}

The parameters of the STTW robot are listed in Table \ref{tab:physical-params}. The nonlinear MPC was developed based on the Control Toolbox library \cite{giftthaler2018control} and deployed on a Raspberry Pi 4B module. The MPC optimization thread \eqref{eq:MPC-NOCP} runs at 100 Hz and the feedback controller \eqref{eq:MPC-controller} runs at 200 Hz.
The rear torque command is sent directly to the rear motor, while the steering velocity command is sent to the steering joint, where a low-level proportional-integral (PI) controller for velocity tracking runs at 1000 Hz.

\subsubsection{Comparison Setting}
In the experiments, to verify the EABC's capability of eliminating slow-varying disturbances, we compare the EABC-MPC presented in this article with three controllers. The first is based on the ECBC scheme proposed in \cite{wang2023tieequilibrium} (ECBC-MPC). The second is an integrator-based MPC \cite{incremona2018ModelPredictive} (IMPC), where the state is augmented by incorporating the integral of the tracking error. The last is a simple MPC controller without a disturbance rejection mechanism (MPC).

\subsection{Track Stand on an Inclined Plane} \label{subsec:adaptation_exp_inclined_plane}

In this experimental scenario, the robot performs the track stand maneuver on a $\text{1m}\times\text{1m}$ board with a varying inclined angle, as depicted in Fig.~\ref{fig:exp-inclinedplane-entire}(b), demonstrating the rejection capability to a varying external disturbance. Initially, the board is placed horizontally. Once the robot is stabilized, the board is tilted longitudinally or laterally, thus applying longitudinal or lateral disturbances to the robot. The angle is gradually increased until the robot loses balance.

Fig.~\ref{fig:exp-inclinedplane-entire}(a) shows the experimental results under the four controllers, and Fig.~\ref{fig:exp-inclinedplane-entire}(c) presents the comparison of the maximum angles and position tracking error of the robot during the experiments. It should be noted that since the inclination angle of the board cannot be directly measured, Fig.~\ref{fig:exp-inclinedplane-entire}(c) presents the pitch and roll angles measured by the robot's inertial measurement unit (IMU). The pitch angle is approximately equal to the longitudinal inclination angle of the board, and the roll angle is positively correlated with the lateral inclination angle. Thus, such angles still characterize the robot's disturbance rejection capabilities under different controllers. In Fig.~\ref{fig:exp-inclinedplane-entire}(c), for the roll/pitch angle, larger values indicate stronger disturbance rejection capabilities; for rear position errors, smaller values represent higher tracking accuracy. The best values in Fig.~\ref{fig:exp-inclinedplane-entire}(c) are highlighted in bold and surrounded by boxes. 

It is clear that, both in the laterally inclined and longitudinally inclined scenarios, both the disturbance rejection capability and the tracking accuracy of the EABC-MPC outperform those under the ECBC-MPC, IMPC and simple MPC. The EABC-MPC controller can maintain the balance of the robot under a maximum roll disturbance of $2.81^\circ$ and a maximum pitch disturbance of $13.31^\circ$ with a position tracking error smaller than $0.05 \text{ m}$. It is remarkable that in the laterally inclined plane, the performance of the EABC method far exceeds that of the other methods. 

\subsection{Rear Position Tracking} \label{subsec:adaptation-exp-rear-tracking}

In this experiment, the robot tracks a reference command for the rear wheel position while performing the track stand, as shown in Fig.~\ref{fig:exp-postrack-entire}(b), aiming to showcase the tracking performance of a varying reference. The initial position is $0\text{ m}$, and then every $16\text{ s}$, the command switches to $0.15\text{ m}$, $0\text{ m}$, $-0.15\text{ m}$, and returns to $0\text{ m}$, in sequence. To avoid command jumps, a first-order low-pass filter is used to smooth the command during switches. The experiments are carried out with the EABC-MPC, ECBC-MPC, IMPC and simple MPC controllers, five times for each controller. The mean value and the $1\text{-}\sigma$ region of the five sets of experimental data are plotted in Fig.~\ref{fig:exp-postrack-entire}(a), represented by solid lines and shaded areas with light colors, respectively. Additionally, RMSE (Root Mean Square Error) and MAE (Mean Absolute Error) are calculated for position/steering tracking, for each set of data, and for every controller. Then the mean and standard deviation of the RMSEs/MAEs for the five sets of data are calculated. These statistics are illustrated in Fig.~\ref{fig:exp-postrack-entire}(c), where the main bar value represents the mean of the five RMSEs/MAEs, and the error bar indicates its standard deviation.

Through Fig.~\ref{fig:exp-postrack-entire}(a) and Fig.~\ref{fig:exp-postrack-entire}(c), it can be observed that the EABC-MPC controller achieves accurate tracking in both the rear position (RMSE: $0.02$ m and MAE: $0.01$ m) and the steering angle (RMSE: $1.08^\circ$ and MAE: $0.84^\circ$). The tracking accuracy of EABC-MPC significantly outperforms other methods by orders of magnitude. Specifically, considering the MAEs of position tracking, the values for ECBC ($0.04$), IMPC ($0.03$) and simple MPC ($0.29$) are $4$ times, $3$ times and $29$ times higher than the one of EABC, respectively. Furthermore, five experiments with EABC-MPC yield a negligible standard deviation of MAE ($6 \times 10^{-4}$), while the values for ECBC, IMPC, and MPC methods are $\mathrm{7 \times 10^{-3}}$, $\mathrm{4 \times 10^{-3}}$ and $\mathrm{6 \times 10^{-3}}$, respectively. The small standard deviation of EABC-MPC demonstrates its high experimental repeatability.

\section{Conclusion}\label{sec:conclusion}
In this article, we answer two questions related to stationary balance control of STTW robots. For the first question regarding the stationary balancing mechanism, we propose utilizing the track stand maneuver and provide an intuitive physical interpretation of this motion. For the second question concerning the rejection of slow-varying disturbances, we introduce a general EABC scheme that effectively handles both matched and mismatched disturbances while ensuring zero steady-state tracking error. The track stand experiments on an inclined plane validate the capability of EABC-MPC to reject varying external disturbances, while the rear position tracking experiments demonstrate the accurate tracking of a varying reference. The experimental results demonstrate that our method significantly outperforms ECBC-MPC, IMPC, and simple MPC in terms of tracking errors by orders of magnitude. In the future, we plan to apply the EABC scheme to other stunt control tasks, such as wheelie, stoppie, and drifting, to further enhance the practicality of STTW robots.

{\appendices
\section*{Appendix}
\subsection{Modeling of the STTW Robot} \label{sec:appendix-model}

In Fig.~\ref{fig:model_schematic_tbview}a, $O\text{-}XYZ$ is the world frame; $o\text{-}xyz$ is the frame with its origin fixed on $P_4^\prime$, its $x$ axis fixed on the robot and its $z$ axis aligned with the $Z$ axis of $O\text{-}XYZ$. The projection of the steering angle on the horizontal plane $\delta_p$ satisfies
\begin{equation}\label{eq:delta_p}
    \tan{\delta_p} \cos{\varphi} = \tan{\delta} \cos{\lambda}.
\end{equation}
The trail length $c_f$ can be expressed as
\begin{equation}
\begin{aligned}
    c_f \approx c + r \tan{\lambda} (\cos{\delta} - 1).
\end{aligned}
\end{equation}
The deviation angle $\gamma$ satisfies 
\begin{equation}
    \gamma \approx \sin \gamma = \frac{c_f}{b} \sin \delta_p.
\end{equation}
The rear wheel turning radius is given by
\begin{equation}\label{eq:R_r}
    R_r = \frac{b+c-c_f / \cos{\delta_p}}{\tan{\delta_p}} \approx \frac{b}{\tan{\delta_p}}.
\end{equation}

The translational velocity and angular velocity of frame $o\text{-}xyz$ relative to the world frame are given by $\bm{v}_o = [v_r, \omega a, 0]^T$ and $\bm{\omega}=[0,0,\omega]^T$, respectively, expressed in frame $o\text{-}xyz$. The acceleration of $o\text{-}xyz$ satisfies 
\begin{equation}
\begin{aligned}\label{eq:a_o}
    \bm{a}_o = \dot{\bm{v}}_o + \bm{\omega} \times \bm{v}_o. 
\end{aligned}
\end{equation}
The roll motion is influenced only by the lateral component of $\bm{a}_o$, which can be obtained through \eqref{eq:delta_p}-\eqref{eq:a_o} as 
\begin{equation} \label{eq:a_oy}
\begin{aligned}
    a_{o,y} &= \omega v_{o,x} + \dot{v}_{c,y} = \omega v_r + \dot{\omega} a \\
    &=\frac{v_{r}^{2}\tan \delta \cos \lambda}{b\cos \varphi}\,\,+\frac{a\cos \!\:\lambda}{b\cos \varphi}\left( \dot{v}_r\tan \!\:\delta +\frac{v_r\dot{\delta}}{\cos ^2\!\:\delta} \right) \\
    &+\frac{av_r\tan \delta \cos \lambda \tan \varphi}{b \cos \varphi}\dot{\varphi}.
\end{aligned}
\end{equation}
According to Fig.~\ref{fig:model_schematic_tbview}b, consider the roll motion around the $x$ axis in the reference frame $o\text{-}xyz$, the robot is subject to gravity $F_g$, inertial force $F_i$ generated by the translational motion of the frame, centrifugal force $F_c$ caused by the rotation motion of the frame, and normal force $N$. These forces satisfy
\begin{equation*}
\begin{aligned}
    F_g &= mg,\ F_i = m a_{o,y}, \ F_c = m\omega^2 h \sin{\varphi}, \\
    N   &= mg - mh\ddot{\varphi}\sin{\varphi}.
\end{aligned}
\end{equation*}
Consequently, the equation of the roll motion is obtained as 
\begin{equation}\label{eq:euler}
    (I_t+mh^2) \ddot{\varphi} = F_g h\sin \varphi + F_c h \cos \varphi + N \gamma a - F_i h \cos \varphi + d_\varphi,
\end{equation}
where $d_\varphi$ is the external disturbance torque applied on the roll channel. 

The acceleration of the rear wheel is governed by 
\begin{align}
    F + d_{r,1}&= m \dot{v}_r \label{eq:rear_acc_trans}\\
    \tau_r + d_{r,2} - F r &= I_r \dot{v}_r / r, \label{eq:rear_acc_rot}
\end{align}
where $d_{r,1}$ is the disturbance torque caused by ground inclination or friction between front-wheel and ground, and $d_{r,2}$ is the disturbance torque directly acted on the rear-wheel motor, such as the friction in the transmission system.

Collecting \eqref{eq:delta_p}-\eqref{eq:rear_acc_rot}, and by selecting the state as $\bm{x} = [s, v_r, \delta, \varphi, \dot{\varphi}]^T$, the control input as $\bm{u} = [\tau_r, \dot{\delta}]^T$, and the disturbance as $\bm{d} = [d_{r1}r + d_{r2}, d_\varphi]^T$, the whole model of the system can be represented as $\dot{\bm{x}} = \bm{f}_{sys}(\bm{x},\bm{u},\bm{d})$, with
\begin{equation}\label{eq:f_sys}
\bm{f}_{sys} = \left[ \begin{aligned}
	&v_r \\
	&\frac{r}{I_r + m r^2} (\tau_r + d_r)\\
	&\dot{\delta}\\
	&\dot{\varphi}\\
	&\frac{1}{mh^2+I_b}\left[ mh c_\varphi c_\lambda \left( \frac{1}{b}t_\delta v_r^2+\frac{a}{b}t_\delta \frac{\tau_r+\tau_d}{I_r + m r^2}  \right.\right.\\
	&\quad \quad \left. \left. +\frac{a}{b}\frac{1}{c^2_\delta}\dot{\delta} v_r \right) - mg\frac{ac}{b}c_\lambda \delta +mgh s_\varphi + d_\varphi \right]
\end{aligned} \right],
\end{equation}
where $s_x$, $c_x$ and $t_x$ are the abbreviations of $\sin x$, $\cos x$ and $\tan x$, respectively.

The coefficients $\beta_1, \cdots, \beta_5$ for \eqref{eq:fx_Gux}\eqref{eq:Gdx} are given by
\begin{align*}
    &\begin{aligned}
     \beta_1(\bm{x}) = 
        \frac{1}{mh^2+I_b} 
        &\left[ mh c_\varphi c_\lambda \left( \frac{1}{b}t_\delta v_r^2  +\frac{a}{b}\frac{1}{c^2_\delta}\dot{\delta} v_r \right)\right. \\
	  & \quad \left.- mg\frac{ac}{b}c_\lambda \delta +mgh s_\varphi \right],
     \end{aligned}\\
    &\beta_2 = \frac{r}{I_r + m r^2},\ 
     \beta_3(\bm{x}) = \frac{ amh c_\varphi c_\lambda t_\delta}{b(mh^2+I_b)(I_r + m r^2)},\\
    &\beta_4(\bm{x}) = \frac{ amh c_\varphi c_\lambda v_r}{b(mh^2+I_b)c_\delta^2},\ 
    \beta_5 = \frac{1}{mh^2+I_b}.
\end{align*}

}

\vfill

\end{document}